# Factorized Multi-Modal Topic Model

**Seppo Virtanen**[1], **Yangqing Jia**[2], **Arto Klami**[1], **Trevor Darrell**[2]
[1]Helsinki Institute for Information Technology HIIT
Department of Information and Compute Science, Aalto University
[2]UC Berkeley EECS and ICSI

## Abstract

Multi-modal data collections, such as corpora of paired images and text snippets, require analysis methods beyond single-view component and topic models. For continuous observations the current dominant approach is based on extensions of canonical correlation analysis, factorizing the variation into components shared by the different modalities and those private to each of them. For count data, multiple variants of topic models attempting to tie the modalities together have been presented. All of these, however, lack the ability to learn components private to one modality, and consequently will try to force dependencies even between minimally correlating modalities. In this work we combine the two approaches by presenting a novel HDP-based topic model that automatically learns both shared and private topics. The model is shown to be especially useful for querying the contents of one domain given samples of the other.

## 1 INTRODUCTION

Analysis of objects represented by multiple modalities has been an active research direction over the past few years. If the analysis of a single modality is characterized as learning some sort of components that describe the data, the task in analysis of multiple modalities can be summarized as learning components that describe both the variation within each modality but also the variation shared between them (Klami and Kaski, 2008; Jia et al., 2010). The fundamental problem is in learning how to correctly factorize the variation into the shared and private components, so that the components can be intuitively interpreted. For continuous vector-valued samples the problem can be solved efficiently by a structural sparsity assumption (Jia et al., 2010; Virtanen et al., 2011), resulting in an extension of canonical correlation analysis (CCA) that models not only the correlations but also components private to each modality.

One prototypical example of multi-modal analysis is that of modeling collections of images and associated text snippets, such as captions or contents of a web page. When both text and image content can naturally be represented with bag of words -type vectors, the assumptions made by the above methods fail. Instead, such count data calls for topic models such as latent Dirichlet allocation (LDA): several extensions of LDA have been presented for multi-modal setups, including Blei and Jordan (2003); Mimno and McCallum (2008); Salomatin et al. (2009); Yakhnenko and Hovavar (2009); Rasiwasia et al. (2010) and Puttividhya et al. (2011). However, none of these extensions are able to find shared and private topics in the same sense as the CCA-based models do for continuous data. Instead, the models attempt to enforce strong correlation between the modalities, which is a reasonable assumption when analyzing e.g. multi-lingual textual corpora with similar languages but that does not hold for analysis of images associated with free-flowing text. In most cases, the images will contain a considerable amount of information not related to the text snippet, and it is not even guaranteed that the text is related at all to the visual content of the image.

In this work, we introduce a novel topic model that combines the two above lines of work. It builds on the correlated topic models (CTM) by Blei and Lafferty (2007) and Paisley et al. (2011), by modeling correlations between topic allocations and by using a hierarchical Dirichlet process (HDP) formulation for automatically learning the number of the topics. The proposed factorized multi-modal topic model integrates the technical improvements of these single-modality topic models to the multi-modal application, and in particular automatically learns to make some topics specific to each of the modalities, implementing the factorization idea of Klami and Kaski (2008) and Jia et al. (2010) used for continuous data. The component selection plays a crucial role in implementing this property, implying that the HDP-based technique for automatically selecting the complexity is even more important for factorized multi-modal models than it would be a for a regular topic model.

The primary advantage of the new model is that is does not enforce correlations between the modalities, like the earlier multi-modal topic models do, but instead factorizes the variation into interpretable topics describing shared and private structure. The model is very flexible and does not enforce any particular factorization structure, but instead learns it from the data. For example, the model can completely ignore the shared topics in case the modalities are independent or find almost solely shared topics when they are strongly correlated. In this work we demonstrate the model in analyzing modalities that have only weak relationships, a scenario for which the previous models would not work. In particular, we analyze a collection of Wikipedia pages that consist of images and the whole text on the page. Such a collection has relatively low between-modality correlation and in particular includes considerable amount of text that is not related to the image at all, necessitating topics private to the text modality. The proposed model is shown to clearly outperform alternative HDP-based topic models as well as correspondence LDA (Blei and Jordan, 2003) in the task of inferring the contents of a missing modality.

## 2 BACKGROUND: TOPIC MODELS

To briefly summarize the topic models and to introduce the notation used in the paper, we describe the standard topic model of Latent Dirichlet Allocation (LDA) (Blei et al., 2003) through its generative process. We assume that words occurring in a document are drawn from $K$ topics. Each topic specifies a multinomial probability distribution over the vocabulary, parameterized through $\boldsymbol{\eta}_k$ drawn from the Dirichlet distribution $\mathrm{Dir}(\gamma\mathbf{1})$, and the topic proportions are multinomial with parameters $\boldsymbol{\theta} \sim \mathrm{Dir}(\nu\mathbf{1})$. The documents are generated by repeatedly sampling a topic indicator $z \sim \mathrm{Multi}(\boldsymbol{\theta})$ and then drawing a word from the corresponding topic as $x \sim \mathrm{Multi}(\boldsymbol{\eta}_z)$.

We will also heavily depend on the concept of correlated topic models (CTM) (Blei and Lafferty, 2007). In the standard LDA the topic proportions $\boldsymbol{\theta}$ drawn from the Dirichlet distribution become independent except for weak negative correlation stemming from the normalization constraint. CTM replaces this choice by logistic normal distribution, first drawing an auxiliary variable from a Gaussian distribution $\boldsymbol{\xi} \sim \mathrm{N}(\boldsymbol{\mu}, \boldsymbol{\Sigma})$ and specifying the topic distribution as $\boldsymbol{\theta} \propto \exp(\boldsymbol{\xi})$. The topics become correlated when $\boldsymbol{\Sigma}$ is not diagonal, and empirical experiments show increased predictive accuracy.

Finally, our model will be formulated through a hierarchical Dirichlet process (HDP) formulation (Teh et al., 2006), to enable automatic choice of the number of topics. As mentioned in the introduction, the choice is even more critical for multi-modal models, since we will have several sets of topics instead of just a single one; specifying the complexity for all of those in advance would not be feasible. Our model will use elements from the recently introduced Discrete Infinite Logistic Normal (DILN) model by Paisley et al. (2011), which incorporates HDP into CTM. The key idea of DILN is that the topic distributions $\boldsymbol{\theta}$ are made sparse by multiplying the $\exp(\boldsymbol{\xi})$ by sparse topic-selection terms. The topic distribution is given by $\boldsymbol{\theta}_k \propto \mathrm{Gamma}(\beta\mathbf{p}_k, \exp(-\boldsymbol{\xi}_k))$, where both $\beta$ and $\mathbf{p}_k$ come from a stick-breaking process: $\beta$ is the second level consentration parameter, and $\mathbf{p}_k = V_k \prod_{i=1}^{k-1}(1 - V_i)$, where $V_k \sim \mathrm{Beta}(1, \alpha)$ with $\alpha$ as the first level concentration parameter. The expected value of $\boldsymbol{\theta}_k$ is proportional to $\beta\mathbf{p}_k \exp(\boldsymbol{\xi}_k)$, illustrating the way the different parameters influence the topic weights. For any finite data collection, $\mathbf{p}_k > 0$ only for a finite subset of topics and hence the model automatically selects the number of topics.

## 3 FACTORIZED MULTI-MODAL TOPIC MODEL

Consider a collection of documents each containing $M$ weakly correlated modalities, where each modality has its own vocabulary. In the application of this paper the two vocabularies are textual and visual words collected from Wikipedia pages with text and a single image (though the model would directly generalize to multiple images). We introduce a novel multi-modal topic model that can be used to learn dependencies between these modalities, enabling e.g. predicting the textual content associated with a novel image. The problem is made particularly challenging by the weak relationship between the modalities; several of the documents will contain large amounts of text not related to the image content.

For modeling the data, we will use $M$ separate vocabularies, so that words (or visual words) for each modality are drawn from separate dictionaries $\boldsymbol{\eta}^{(m)}$ specific to each view $m$. The topic proportions $\boldsymbol{\theta}^{(m)}$ will also be specific to each modality, whereas the actual words are sampled independently for each modal-ity given the topic proportions. The essential modeling question is then how the topic proportions are tied with each other, in order to achieve the factorization into shared and private topics. In brief, we will do this by (i) modeling dependencies between topics both within and across modalities and (ii) automatically selecting the number of topics for each type (shared or private to any of the modalities).

The topic proportions $\boldsymbol{\theta}^{(m)}$ are made dependent by introducing auxiliary variables $\boldsymbol{\xi}^{(m)}$, denoting by $\boldsymbol{\xi} = (\boldsymbol{\xi}^{(1)}, ..., \boldsymbol{\xi}^{(M)})$ the concatenation of them, and using the CTM prior $\boldsymbol{\xi} \sim \mathrm{N}(\boldsymbol{\mu}, \boldsymbol{\Sigma})$. This part of the model corresponds to the 'multi-field CTM with different topic sets' by Salomatin et al. (2009), and the different blocks in $\boldsymbol{\Sigma}$ describe different types of dependencies between the topic proportions. In particular, the blocks around the diagonal describe dependencies between the topic proportions of each modality, whereas the off-diagonal blocks describe dependencies in topic proportions between the modalities.

Having a CTM for the joint topic distribution is not yet sufficient for separating the shared topics from private ones, since we can only control the correlation between the topic proportions. A large correlation between two topics for different modalities would imply that it is shared, but lack of correlation (that is, $\boldsymbol{\Sigma}_{kl} = 0$) would not make either component private. Instead, the weights would simply be determined independently. To create separate sets of shared and private topics we need to be able to switch some of the topics off in one or more of the modalities, similarly to how Jia et al. (2010) and Virtanen et al. (2011) switch off components to make the same distinction in continuous data models. In the case of multi-field CTM this could only be done by driving $\boldsymbol{\mu}_k$ (the mean of the Gaussian prior for $\boldsymbol{\xi}_k$) towards minus infinity, which is not encouraged by the model and is difficult to achieve with mean-field updates.

We implement the shared/private choice by separate HDPs, one for each modality, switching a subset of topics off for each modality separately by a mechanism similar to how the single-view DILN model (Paisley et al., 2011) selects the topics. We introduce $\beta^{(m)}$ and $\mathbf{p}^{(m)}$ for each modality $m = 1, ..., M$, and draw them from separate HDPs, resulting in $\boldsymbol{\theta}^{(m)} \propto \mathrm{Gamma}(\beta^{(m)}\mathbf{p}^{(m)}, \exp(-\boldsymbol{\xi}^{(m)}))$ as the final topic proportions. The topic distributions are still shared through $\boldsymbol{\xi}^{(m)}$ that were drawn from a single high-dimensional Gaussian, but for each modality the stick weights $\mathbf{p}^{(m)}$ select different subsets of topics to be switched off. In the end, a finite number of topics remain for each modality, and the private topics can be identified as ones that have non-zero weight for one modality and are not correlated with topics active in other modalities.

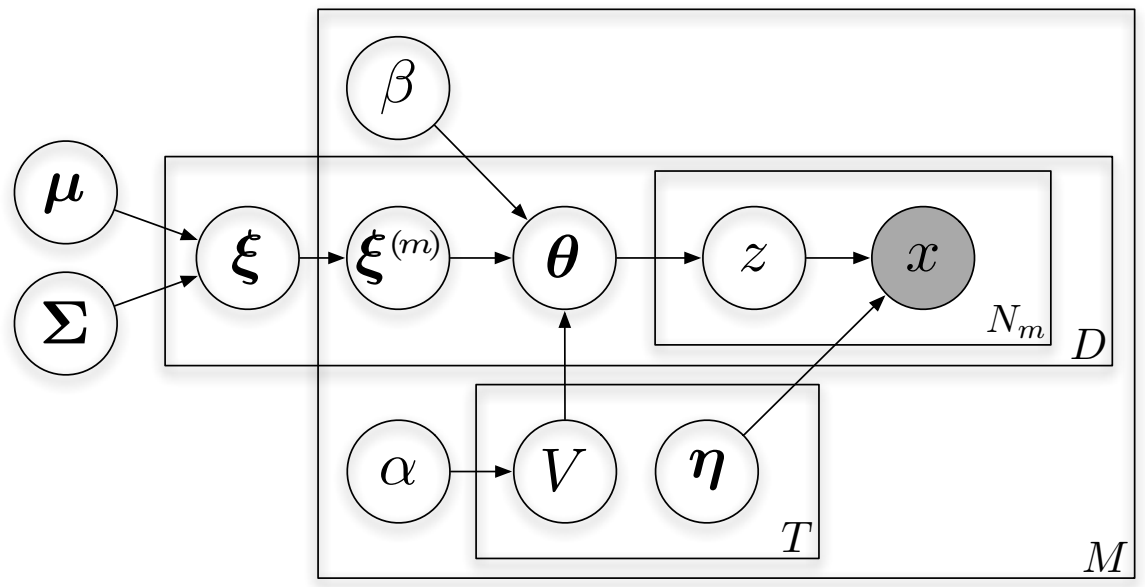

Figure 1: A graphical representation of the factorized multi-modal topic model. The data has $D$ documents described by $M$ modalities. For each modality, the words $\mathbf{x}^{(m)}$ are drawn from dictionary specific to that modality, according to topic proportions $\boldsymbol{\theta}^{(m)}$ also specific to the modality. The topic proportions are generated by logistic transformation of latent variables $\boldsymbol{\xi}^{(m)}$ that model the correlations between the topics both within and across modalities, followed by topic selection with a HDP (denoted by $V$ and $\beta$ in the plate; see text for details) for each modality. As a result, the model learns both topics modeling correlations between the modalities as well as topics private to each modality.

The final generative model motivated by the above discussion results in a collection of $M$ correlated BOW data sets $\mathbf{X}^{(m)}$, generated as follows (see Figure 1 for graphical representation). For the whole collection we:

- create a dictionary of $T^{(m)}$ topics for each modality by drawing $\boldsymbol{\eta}_k^{(m)} \sim Dir(\gamma^{(m)}\mathbf{1})$ for $k = 1, .., T^{(m)}$
- draw the parameters $\alpha^{(m)}, \beta^{(m)}, V^{(m)}$ of the DILN distribution for each modality from the stick-breaking formulation and construct $\mathbf{p}^{(m)}$.

For each document $d$ we then draw $\boldsymbol{\xi} \sim \mathrm{N}(\boldsymbol{\mu}, \boldsymbol{\Sigma})$ and partition it into the different modalities as $\boldsymbol{\xi} = (\boldsymbol{\xi}^{(1)}, ..., \boldsymbol{\xi}^{(M)})$. For each modality, we then generate the words independently as follows:

- form the topic proportion by drawing $Y_k^{(m)} \sim \mathrm{Gamma}(\beta^{(m)}\mathbf{p}_k^{(m)}, \exp(-\boldsymbol{\xi}_k^{(m)}))$ and set $\boldsymbol{\theta}_k^{(m)} = \frac{Y_k^{(m)}}{\sum_{i=1}^{T^{(m)}} Y_i^{(m)}}$
- draw $N^{(m)}$ words by choosing a topic $z \sim \mathrm{Multi}(\boldsymbol{\theta}^{(m)})$ and drawing a word $x \sim \mathrm{Multi}(\boldsymbol{\eta}_z^{(m)})$

### 3.1 INFERENCE

For learning the model parameters we use a truncated variational approximation following closely the algorithm given by Paisley et al. (2011), the main difference being that we have $M$ separate sets of $\boldsymbol{\eta}$, $\beta$ and $\mathbf{p}$, one for each modality. The above generative process is truncated by setting $V^{(m)}_{T^{(m)}} = 1$, forcing the stick lengths beyond the truncation level $T^{(m)}$ to be zero, and the resulting factorized approximation is given by

$$Q = \prod_{m=1}^{M} \prod_{d=1}^{D} \prod_{n_m=1}^{N_m} \prod_{k=1}^{T} q(z^{(m)}_{dn_mk}) q(Y^{(m)}_{dn_mk}) q(\xi^{(m)}_{dk}) q(\boldsymbol{\eta}^{(m)}_k)$$
$$q(V^{(m)}_k) q(\alpha_m) q(\beta_m) q(\boldsymbol{\mu}) q(\boldsymbol{\Sigma}),$$

where to simplify notation we assume $T_m = T \; \forall \; m$. The algorithm proceeds by updating each factor in turn while keeping the others fixed, using either gradient ascent or analytic solution for maximizing the lower bound of the approximation for each of the terms (see Paisley et al. (2011) for details).

The main difference in the algorithms comes from updating $\boldsymbol{\xi}$, since in our case it goes over $M$ sets of topics instead of just one, yet the activities within each set are governed by separate HDPs. We use a diagonal Gaussian factor $q(\boldsymbol{\xi}) = \mathrm{N}(\tilde{\boldsymbol{\xi}}, \mathrm{diag}(\tilde{\mathbf{v}}))$, where $\tilde{\mathbf{v}}$ denotes the variances of the dimensions, and use gradient ascent for jointly updating the parameters. To simplify notation we use $\boldsymbol{\xi}$ and $\mathbf{v}$ to denote the expectation and variance of the factorial distribution. The relevant part of the lower bound is

$$\begin{aligned} \mathcal{L}_{\boldsymbol{\xi},\mathbf{v}} = & -\sum_{m=1}^{M} \beta^{(m)} \mathbf{p}^{(m)^T} \boldsymbol{\xi}^{(m)} \\ & -\sum_{m=1}^{M} \mathbb{E}[\boldsymbol{\theta}^{(m)}]^T \mathbb{E}[\exp(-\boldsymbol{\xi}^{(m)})] \\ & - (\boldsymbol{\xi} - \boldsymbol{\mu})^T \boldsymbol{\Sigma}^{-1} (\boldsymbol{\xi} - \boldsymbol{\mu})/2 \\ & - \mathrm{diag}(\boldsymbol{\Sigma}^{-1})^T \mathbf{v}/2 + \log(\mathbf{v})^T \mathbf{1}/2. \end{aligned} \quad (1)$$

Here $\boldsymbol{\Sigma}^{-1}$ couples the separate $\boldsymbol{\xi}^{(m)}$ terms in the partial derivatives as

$$\begin{aligned} \frac{\partial \mathcal{L}_{\boldsymbol{\xi},\mathbf{v}}}{\partial \boldsymbol{\xi}^{(m)}} = & -\beta^{(m)} \mathbf{p}^{(m)} + \mathbb{E}[\boldsymbol{\theta}^{(m)}] \mathbb{E}[\exp(-\boldsymbol{\xi}^{(m)})] \\ & - (\boldsymbol{\Sigma}^{-1})_{m,m} (\boldsymbol{\xi}^{(m)} - \boldsymbol{\mu}^{(m)}) \\ & - \sum_{j \neq m} (\boldsymbol{\Sigma}^{-1})_{m,j} (\boldsymbol{\xi}^{(j)} - \boldsymbol{\mu}^{(j)}), \end{aligned}$$

with $(\boldsymbol{\Sigma}^{-1})_{i,j}$ denoting a block of $\boldsymbol{\Sigma}^{-1}$ corresponding to modalities $i$ and $j$. The inverse of $\boldsymbol{\Sigma}$ remains constant during the gradient descent, and hence only needs to be evaluated once for every time the factor $q(\boldsymbol{\xi})$ is updated.

We use maximum marginal likelihood to update $\boldsymbol{\mu}$ and $\boldsymbol{\Sigma}$ resulting in closed form updates

$$\boldsymbol{\mu} = \frac{1}{D} \sum_{d=1}^{D} \boldsymbol{\xi}_d$$
$$\boldsymbol{\Sigma} = \sum_{d=1}^{D} \big( (\boldsymbol{\xi}_d - \boldsymbol{\mu})(\boldsymbol{\xi}_d - \boldsymbol{\mu})^T + \mathrm{diag}(\mathbf{v}_d) \big)/D.$$

### 3.2 PREDICTION

The model structure is well suited for prediction tasks, where the task is to infer missing modalities for a new document given that one of them is observed (e.g. infer the caption given the image content). This is because the correlations between the topic proportions provide a direct link between the modalities, and the private topics explain away all the variation that is not useful for predictions.

Here we present the details of the prediction for the special case with just one observed modality ($j$) and one missing modality ($i$). Given the observed data we first infer the topic proportions $\hat{\boldsymbol{\theta}}^{(j)}$ and then auxiliary variable $\hat{\boldsymbol{\xi}}^{(j)}$ by maximizing a cost similar to (1), but only using the newly inferred topic proportions of the observed modality and the corresponding part of $\boldsymbol{\Sigma}$. As $\hat{\boldsymbol{\xi}}$ comes from a Gaussian distribution we can infer $\hat{\boldsymbol{\xi}}^{(i)}$ given $\hat{\boldsymbol{\xi}}^{(j)}$ with the standard conditional expectation as

$$\begin{aligned} \hat{\boldsymbol{\xi}}^{(i)} &= \boldsymbol{\mu}^{(i)} + \boldsymbol{\Sigma}_{i,j} \boldsymbol{\Sigma}^{-1}_{j,j} (\hat{\boldsymbol{\xi}}^{(j)} - \boldsymbol{\mu}^{(j)}) \\ &= \boldsymbol{\mu}^{(i)} + \mathbf{W} (\hat{\boldsymbol{\xi}}^{(j)} - \boldsymbol{\mu}^{(j)}). \end{aligned} \quad (2)$$

Here $\mathbf{W}$ involves the corresponding part of the between-topic covariance matrix $\boldsymbol{\Sigma}$ as indicated above, and can be seen as a projection matrix transforming the components of one modality to another. Finally, the newly estimated $\hat{\boldsymbol{\xi}}^{(i)}$ for the missing views is converted back to the expecteed topic proportion $\hat{\boldsymbol{\theta}}$ by exponentiation and multiplying with the corresponding stick lengths $\mathbf{p}^{(i)}$.

### 3.3 SHARED AND PRIVATE TOPICS

The key novelty of the model is its capability to learn both topics that are shared and that are private to each modality, without needing to specify them in advance. Since the way these topics appear is by no means transparent in the above formulation, we will here discuss the property in more detail. In brief, the distinct nature for the topics comes from an interplay of the correlations between the topics of different modalities and the HDP procedure that turns some of the topics off for each modality. In particular, neither of these properties alone would be sufficient.

As mentioned already in Section 3, merely having separate $\boldsymbol{\xi}^{(m)}$ drawn from a single Gaussian is not sufficient for finding private topics. At best, the correlation structure can specify that the weights will be independent for the modalities. Next we explain how the other key element of the model, separate selection of active topics for each modality, is not sufficient alone either. We do that by considering a special case of the model that assumes equal $\boldsymbol{\xi} = \boldsymbol{\xi}^{(m)}$ for all views but has separate stick-breaking processes switching some of the topics off for each of the views. We call this alternative model mmDILN, due to the fact how it implementes multi-modal LDA of Blei and Jordan (2003) with DILN-style component selection.

Intuitively, mmDILN model could find private topics simply by setting $\mathbf{p}_k^{(m)}$ to small value for topics that are not needed in that modality. However, it cannot make correct predictions from one modality to another, and hence fails in achieving one of the primary goals for shared-private factorizations. If $\mathbf{p}_k^{(m)}$ is small then the model has no information for inferring $\boldsymbol{\xi}_k$ from that view, and hence also all other elements $\boldsymbol{\xi}_l$ that correlate with $\boldsymbol{\xi}_k$ will be incorrect. If $\boldsymbol{\xi}_k$ was an important topic for the other view, the predictions will be severely biased. Our model avoids this issue by having the separate $\boldsymbol{\xi}^{(m)}$ parameters, leading to correct across-modality predictions as described in the previous section. In the experimental section we will empirically compare the proposed model with mmDILN, demonstrating how mmDILN indeed has very poor predictive accuracy despite modeling the training data almost as well. Hence, even though the structure is in principle sufficient for learning private topics, the model has no practical value as a shared-private factorization.

In order to recognize the nature of each of the topics, we need to look at both the covariance $\boldsymbol{\Sigma}$ between the topic weights and the modality-specific stick weights $\mathbf{p}_k^{(m)}$. Since the topics can be (potentially strongly) correlated both within and across modalities, we can identify private topics only by searching for topics that do not correlate with any topic that would be active in any other modality. In the experiments we demonstrate how the topics can be ranked according to how strongly they are shared with another modality, by inspecting the elements of $\boldsymbol{\Sigma}$.

# 4 RELATED WORK

In this section we relate the model to other approaches for modeling multi-modal count data.

## 4.1 MULTI-MODAL TOPIC MODELS

The multi-modal extension of LDA (mmLDA) by Blei and Jordan (2003) and its non-parametric version mmHDP by (Yakhnenko and Hovavar, 2009) assume all modalities to share the same topic proportions, and essentially extend LDA only by having separate dictionaries for each modality and generating the words for the domains independently. For many real world data sets the assumption of identical topic proportions is too strong, and the model tries to enforce correlations even when they do not exist. While the assumption may help in picking up topics that would be weak in either modality alone, it makes identifying the true correlations almost impossible.

Such models fail especially when modeling data having strong private topics in one modality. Since the topic proportions are shared, the topic must be present in other modalities as well and becomes associated with a dictionary that merely replicates the overall distribution of the words. Such topics are particularly harmful for prediction tasks. When the dictionary of a topic matches that of the background word distribution, it will be present in every document in that modality. For example, when predicting text from images we could learn to associate politics (a strong topic private to the text modality) with the overall visual word distribution, resulting in all of the predictions including terms from the politics topic.

Salomatin et al. (2009) took a step towards our model with their multi-field CTM. It extends CTM by introducing separate $\boldsymbol{\xi}^{(m)}$ for each modality, similarly to our model. However, as described in the previous section the separate topic proportions are not yet sufficient for separating the shared topics from private ones.

## 4.2 CONDITIONAL TOPIC MODELS

Lots of recent work on multi-modal topic modeling framework has focused on building conditional models, largely for image annotation task. Correspondence LDA (corrLDA) proposed simultaneously to mmLDA in (Blei and Jordan, 2003) is a prominent example, assuming that the image is generated first and the text depends on the image content. Both modalities are assumed to share the same topic weights. While such models are very useful for modeling the conditional relationship, they do not treat the modalities symmetrically as in our model. Recently Puttividhya et al. (2011) proposed an extension of corrLDA, replacing the identical topic distributions with a regression module from image topics to the textual annotation topics. The added flexibility results in better predictive performance, but the model remains a directional one, in contrast to our model that generates all modalities with equal importance. For applications treating only two modalities and having a specific task that makes one of them more important (say, image annotation) the conditional models often work well. However, they do not easily generalize to multiple modalities and are not flexible in terms of the eventual application.

Other conditional models focus on conditioning on meta-data, such as author or link structure (Mimno and McCallum, 2008; Hennig et al., 2012). Such models allow integrating data that are not necessarily in count format, but the same distinction of directional versus generative applies. However, this family of models could be integrated with our solution, incorporating a meta-data link into our multi-modal model. In essence, the choice of whether meta-data is modeled or not is independent of the choice of how many count data modalities the data has.

### 4.3 CANONICAL CORRELATIONS

As described earlier, the model bears close resemblance to how CCA models correlations between continuous data, the similarities being most apparent with the recent re-interpretations of CCA as shared-private factorization (Klami and Kaski, 2008; Jia et al., 2010). The technical details of the solutions are, however, very different as the normalization of topic proportions makes the techniques used for continuous data not feasible for topic models.

Despite the mismatch of data types, CCA can be used for modeling count data as well. The most promising direction would be to apply kernel-CCA, but there are no obvious choices for the kernel function that would directly match the analysis of image-text pairs. As one practical remedy, (Rasiwasia et al., 2010) combined CCA and LDA directly by first estimating a separate LDA model for each modality and then combining the resulting topic proportions with CCA. Our approach does not rely on two separate analysis steps that do not result in directly interpretable private topics.

## 5 EXPERIMENTS AND RESULTS

### 5.1 DATA AND MEASURES

We validate the model on real data collected from Wikipedia[1]. We constructed a data collection with $D = 20,000$ documents, each consisting of a single image represented with 5000 SIFT patches and text (the contents of the whole Wikipedia page) represented with a vocabulary of 7500 most frequent terms, after stopword removal. We make a random 50/50-split into test and train data. To demonstrate the ability of the proposed model to correctly model the relationships between the two modalities, we evaluate the model with conditional perplexity of a missing modality for a new sample:

$$\mathcal{P}_{\text{train}}^{(m)} = \exp\left( - \frac{\sum_{d \in D_{train}} \log p(\mathbf{x}_d^{(m)})}{\sum_{d \in D_{train}} \mathbf{x}_d^{(m)}} \right)$$

$$\mathcal{P}_{\text{test}}^{(i)|(j)} = \exp\left( - \frac{\sum_{d \in D_{test}} \log p(\mathbf{x}_d^{(i)} | \mathbf{x}_d^{(j)})}{\sum_{d \in D_{test}} \mathbf{x}_d^{(i)}} \right),$$

where $\mathbf{x}_d^{(m)}$ denotes concatenation of $N_d^{(m)}$ words. These quantities measure how well the model can relate the visual content to the textual content, corresponding to the document completion task of Wallach et al. (2009) but computed across modalities.

We compare our model to three alternatives representing various kinds of multi-modal topic models: mmDILN (Section 3.3), mmHDP (Section 4.1) and corrLDA (Section 4.2). Both mmDILN and mmHDP are comparable to our model in making automatic topic number selection and modeling both modalities symmetrically. Consequently, the experiments will focus on demonstrating the importance of finding the correct factorization into shared and private topics. The corrLDA is included as an example of a conditional model that gives an alternative approach to solving a similar prediction task. Note that we need to learn two separate corrLDA models, one for predicting text from images and one for the other direction, whereas the other models can do both types of predictions. For corrLDA we use 100 topics (the threshold we used for nonparametric models).

### 5.2 INFERENCE SPEED

First we show that the variational approximation used for inference is efficient. Figure 2 shows how the algorithm converges for both $N = 400$ and $N = 10000$ documents already after some tens of iterations. For both experiments we used a maximum of $T = 100$ topics. The convergence of mmHDP and mmDILN is similar (not shown).

### 5.3 PREDICTING TEXT FROM IMAGES AND VISE VERSA

Figure 3 shows the evaluation for training and test sets for the proposed model and the comparison methods, measured as the perplexity on training data and the conditional perplexity of images given the text and text given the images. The proposed method, which is more flexible than the alternatives, reaches better (lower) perplexity on the training and testing data due to being able to describe both variation not shared by the other modality without needing to introduce noise topics.

[1]Available from `http://www.eecs.berkeley.edu/~jiayq/`

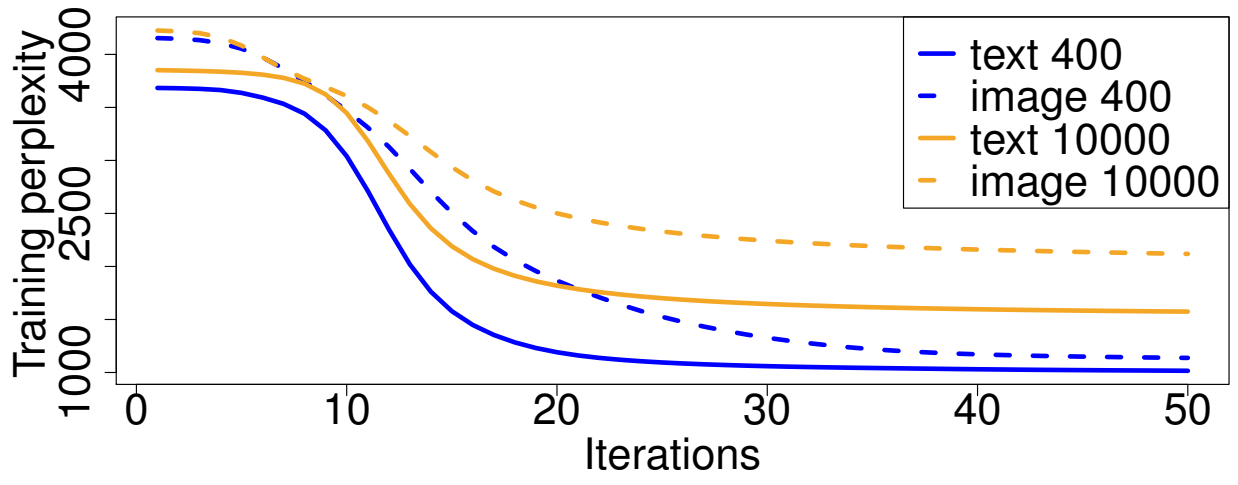

Figure 2: Training perplexity as function of algorithm iterations.

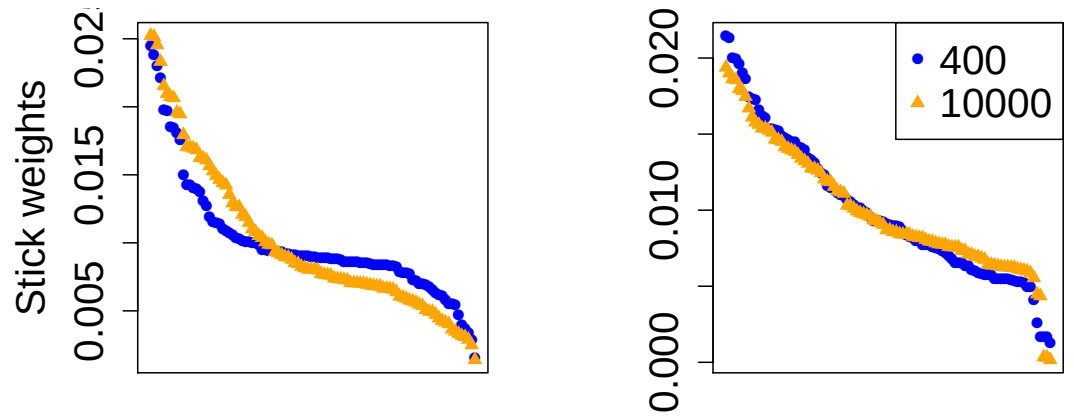

(a) Our model: text (b) Our model: image

Figure 4: Visualization of stick parameter $\mathbf{p}$ of the proposed model for the text modality (a) and the image modality (b) reveals how they are not identical for the two modalities. Both figures show the weights for two models learned with 400 and 10,000 documents, revealing how the distribution is learned fairly accurately already from a small collection.

A notable observation is that the baseline methods perform worse at predicting text from images as the amount of training data increases. This illustrates clearly the fundamental problem in modeling multi-modal collections without separate private topics. Since the text documents are easier to model than the images, the alternative models start to focus more and more on modeling the text when there is large amount of data. The dominant topics start describing the text alone, yet they are also active in the image modality but with a topic that does not contain any information. Given a new image sample, the estimated topic proportions will be arbitrary and hence do not enable meaningful prediction. The proposed model, however, learns to make those textual topics private to the text modality, while capturing weaker correlations between the two modalities with shared topics. The model still cannot predict textual information not correlated with the image content, but it learns correctly not to even attempt that and manages to make accurate predictions for the aspects that are correlated.

### 5.4 SHARED AND PRIVATE TOPICS

To illustrate how the HDP-formulation chooses the topics, we visualize the stick parameters $\mathbf{p}$ in Figure 4. First, we notice that the last sticks have close to zero weight, indicating that the chosen truncation level $T = 100$ is sufficient. More importantly, we see that the weights for the text and image topics are different (the image topics are more spread out), motivating the choice of separate weights for the modalities.

To further understand how the proposed model is able to find both shared and private topics, we explore the nature of the individual topics. Since the SIFT vocabulary is not easily interpretable by visual inspection, we illustrate the property for the textual topics. For each textual topic we measure the amount of correlation between the other modality by inspecting the correlation structure in $\mathbf{\Sigma}$, and then rank the topics according to this measure. This results in a ranked list of the text topics, the first ones being strongly shared by the two modalities while the last ones are private to the text modality.

More specifically, denoting the separate blocks in the covariance matrix as

$$\mathbf{\Sigma} = \begin{pmatrix} \mathbf{\Sigma}_{t,t} & \mathbf{\Sigma}_{t,i} \\ \mathbf{\Sigma}_{i,t} & \mathbf{\Sigma}_{i,i} \end{pmatrix}, \quad (3)$$

we convert it to a correlation matrix, $\mathbf{\Omega}$, threshold small values out (we used a threshold of 0.2) and extract the cross-correlation between textual (rows) and visual topics (columns), to get $\mathbf{\Omega}_{t,i}$. Then for each textual topic we define visual relevance, $\boldsymbol{\rho}_k$, as row mean of absolute values of $\mathbf{\Omega}_{t,i}$, written as $\boldsymbol{\rho} = \frac{1}{T}|(\mathbf{\Omega}_{t,i})|\mathbf{1}$ [2]. This quantity captures general and rich visual combinations that co-occur with the textual topics, and it is worth noticing how the measure is very general: It allows multiple visual topics to correlate with one textual topic (and vise versa), and includes both positive and negative correlations that are typically equally relevant (negative correlation can be seen as absence of a visual component) (See Figure 5 for demonstration).

The textual topics are ranked according to $\boldsymbol{\rho}$ in Figure 6 . There are a few very strong shared topics between text and image modalities, and at the end of the list we have several topics private to the text modality, indicated by zero correlation with the image modality. This matches with the intuition that the full text of a Wikipedia page cannot be mapped to the image content in all cases. Table 1 summarizes the six text topics most strongly correlating with the image modality, as well as six topics that are private to the text modality, revealing very clear interpretations. The most strongly correlating topic covers airplanes, which are known to be easy to recognize from the images due to the distinct shapes and background. The second topic is about maps that also have clear visual correspondence, and the other strongly correlated topics also cover clearly visual concepts like buildings, cars and railroads. The topics private to the text domain, in turn, are about concepts with no clear visual counterpart: economy, politics, history and research. In summary, the model has separated the components nicely into shared and private ones, and provides additional interpretability beyond regular multi-modal topic models.

[2] We also tried using the maximum element instead of the mean; it results in fairly similar ranking.

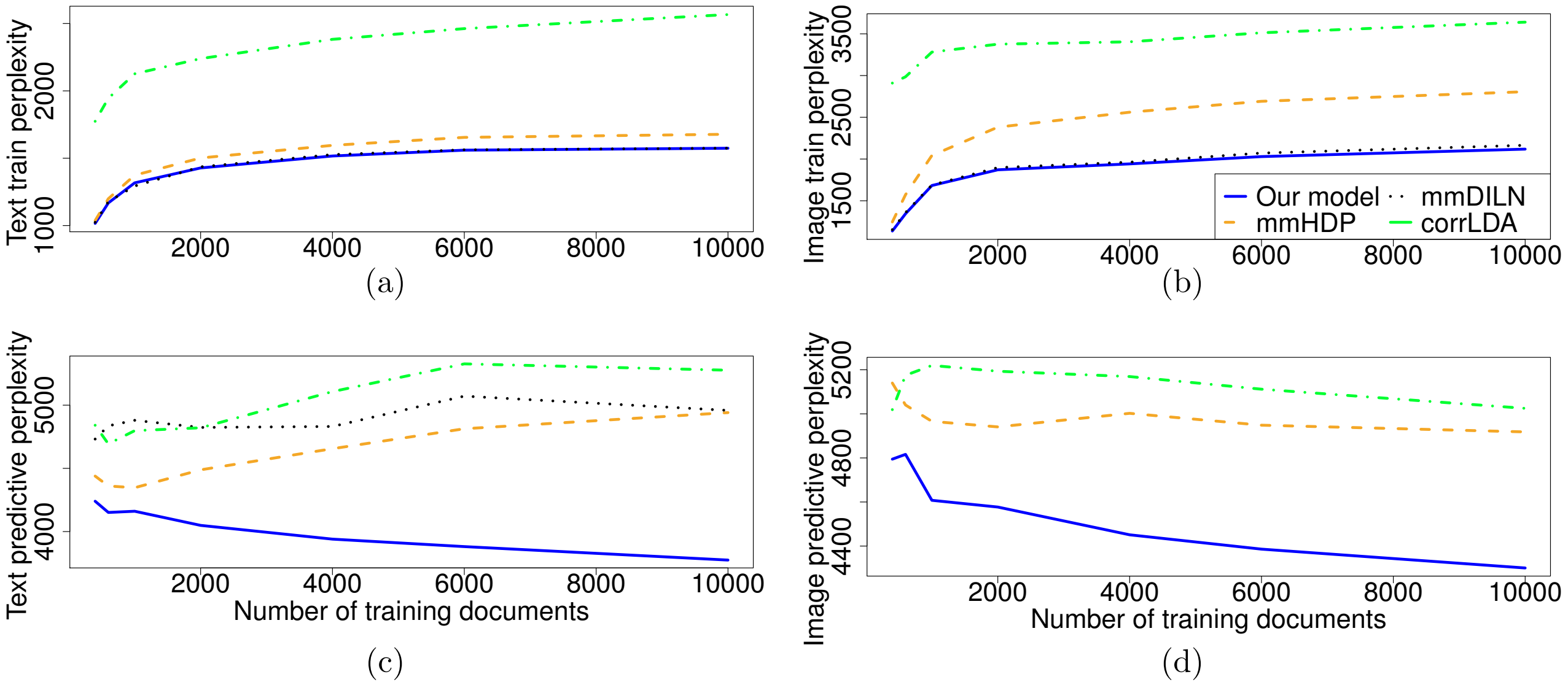

Figure 3: Training and test perplexities (lower is better) for the two modalities. For training data we show the perplexity of modeling the text (a) and images (b) separately. For test data, we show the conditional perplexity of predicting text from images (c) and predicting images from text (d), corresponding to the document completion task used for evaluating topic models. The proposed method outperforms the comparison ones in all respects. The comparison methods mmHDP, mmDILN and corrLDA that are not able to extract topics private to either modality are not able to learn good predictive models, demonstrated especially by the error increasing as a function of training samples in (c). The image prediction perplexity for mmDILN is outside the range depicted in (d), above 5400 for all training set sizes.

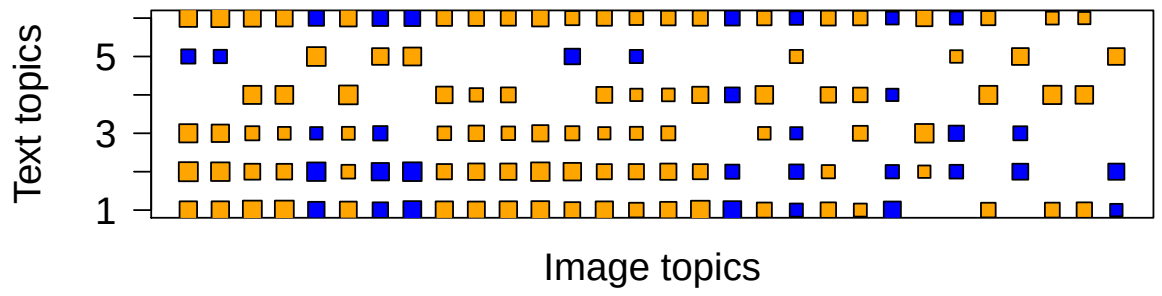

Figure 5: Illustration of part of cross-correlation between text topics and image topics corresponding to subset of $\mathbf{\Omega}_{(t,i)}$, where yellow represents positive correlations, and blue represents negative ones. The size of the boxes corresponds to the absolute value.

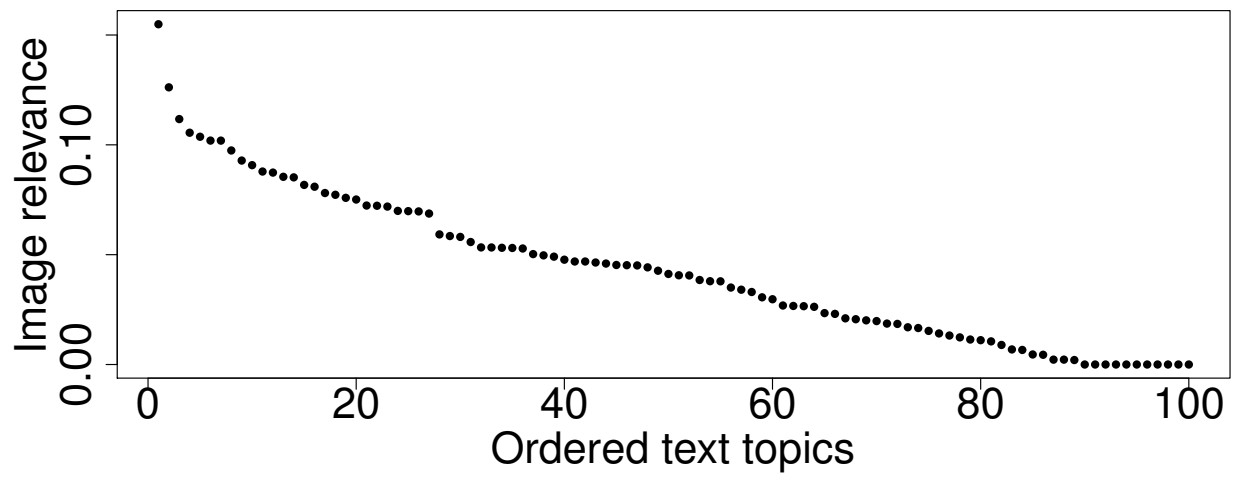

Figure 6: Text topics ordered according to visual relevance $\boldsymbol{\rho}$. We see that there are a few strongly correlating topics, and that the model has found roughly 10 topics that are private to the text domain. Note that such topics may still be important for modeling the whole multi-modal corpus, whereas they do not contribute to the cross-modal information transfer.

# 6 DISCUSSION

Our paper ties together two separate lines of work for analysis of multi-modal data. In particular, we created a novel multi-modal topic model which extends earlier tools for analysis of multi-modal count data by incorporating elements found useful in the continuous-valued case. We explained how learning topics private to each modality is of crucial importance while modeling modalities with potentially weak correlations, and demonstrated empirically how such a property can only be obtained by combining two separate elements: modeling correlations between separate topic weights for each modality, and learning modality-specific indicators switching unnecessary topics off. For implementing these elements we combined state-of-art techniques in topic models, integrating the DILN distribution (Paisley et al., 2011) into a model similar to the multi-field correlated topic model of Salomatin et al. (2009), to create an efficient learning algorithm readily applicable for relatively large document collections.

Table 1: Text topics ranked according to visual relevance, summarized by the words with highest probability. The topic indices match the ranking in Figure 6. The shared topics have clear visual counterparts, whereas the private ones do not relate with any kind of visual content.

| Shared topics |
|---|
| T1 airport flight airlines air international aircraft aviation terminal passengers airline boeing flights airways service airports passenger accident |
| T2 format dms latd dm longm latm longs lats launched mi broken mill sold renamed dec captured rapids class feet coordinates built lake located |
| T3 building house built buildings street hall st century tower houses west designed design castle south north east side main square large end site |
| T4 car engine cars model models ford engines race rear series front racing wheel year driver speed vehicles vehicle production hp motor drive |
| T5 retrieved album song music video released single awards number billboard chart top release mtv songs media love show uk jackson hot albums |
| T6 line railway station rail trains train service lines bus transport services system railways stations built railroad passenger main metro transit |
| **Topics private to the text domain** |
| T95 president washington post united american national states secretary december november september times military dc kennedy press security |
| T96 ottoman turkish turkey kosovo armenian war greek serbia bulgarian serbian government border bulgaria turks forces croatian albanian republic |
| T97 research science development institute university management scientific technology design world national engineering work human international |
| T98 government state national european policy council international states members act union political countries system nations article parliament |
| T99 nuclear weapons anti power protest bomb people protests united protesters government strike peace states march reactor atomic april test |
| T100 economic trade economy world production industry oil million growth development government agricultural market agriculture industrial |

### Acknowledgements

AK and SK were supported by the COIN Finnish Center of Excellence and the FuNeSoMo exchange project. AK was additionally supported by Academy of Finland (decision number 133818) and PASCAL2 European Network of Excellence.